# ArmanTTS single-speaker Persian dataset

1st Mohammd Hasan Shamgholi
MSc Student
School of Computer Engineering
Iran University of Science and Technology
Tehran, Iran
m_shamgholi@comp.iust.ac.ir

2nd Vahid Saeedi
MSc Graduate
School of Computer Engineering
Iran University of Science and Technology
Tehran, Iran
vahid_saeedi@alumni.iust.ac.ir

3rd Javad Peymanfard
PhD Candidate
School of Computer Engineering
Iran University of Science and Technology
Tehran, Iran
javad_peymanfard@comp.iust.ac.ir

4th Leila Alhabib
BSc Student
School of Computer Engineering
Amirkabir University of Technology
Tehran, Iran
alhabib.leila@gmail.com

5th Hossein Zeinali
Assistant Professor
School of Computer Engineering
Amirkabir University of Technology
Tehran, Iran
hzeinali@aut.ac.ir

*Abstract*— **TTS, or text-to-speech, is a complicated process that can be accomplished through appropriate modeling using deep learning methods. In order to implement deep learning models, a suitable dataset is required. Since there is a scarce amount of work done in this field for the Persian language, this paper will introduce the single speaker dataset: ArmanTTS. We compared the characteristics of this dataset with those of various prevalent datasets to prove that ArmanTTS meets the necessary standards for teaching a Persian text-to-speech conversion model. We also combined the Tacotron 2 and HiFi GAN to design a model that can receive phonemes as input, with the output being the corresponding speech. 4.0 value of MOS was obtained from real speech, 3.87 value was obtained by the vocoder prediction and 2.98 value was reached with the synthetic speech generated by the TTS model.**

*Keywords-Text-to-speech dataset; Vocoders; Acoustic models*

## I. INTRODUCTION

By expanding deep learning models, a complicated process like converting text to speech can be done with great accuracy, even generating human-like speech [1]–[4]. In order for deep learning models to perform efficiently, they require ample amounts of suitable data to be able to cover all aspects of the process. There is an abundance of datasets for dealing with English text-to-speech conversion [5]–[7]. However, Persian-text-to-speech[8] is the only available dataset for Persian, showing the lack of research and datasets in this field. In order to improve Persian TTS, We introduce the ArmanTTS as a Farsi text to speech conversion dataset. To evaluate our dataset we used two models on for generating spectrograms and the other for converting the spectrograms to the audio. Table I shows a comparison of ArmanTTS's sample rates, total speaking time, and number of speakers to available prominent datasets in English, German, and Farsi.

The characteristics of the ArmanTTS dataset are listed as follows:

- Sample rate of 22.0g kHz
- Single speaker
- Studio recorded speech
- Model input given is a combination of phonemes and output provided as waveforms
- Audio duration of 9 hours, 12 minutes and 14 seconds
- Average signal to noise ratio of 25dB

In order to evaluate this dataset, we designed a TTS model that receives the phonemes of a sentence as input and creates its corresponding speech. To convert phonemes to acoustic features, the Tacatron 2[9] model is used and to convert acoustic features to waveforms, HiFi-GAN[3] is used. In this paper we review previous datasets and research in the field of TTS in Related Work. Next in Dataset Structure we will look at some features and characteristics of the ArmanTTS dataset. Then, in Experiment, we evaluate the models on the ArmanTTS dataset. The final segment, Conclusion, concludes this paper with an overview and conclusion from our research.

## II. RELATED WORK

TABLE I. TEXT-TO-SPEECH DATASETS

| dataset | Duration (hour) | Sample Rate (kHz) | Multi/single-speaker |
|---|---|---|---|
| LJSpeech | 24 | 22.05 | Single-speaker |
| LibriTTS | 586 | 24 | Multi-speaker |
| HiFi-TTS | 292 | 44.1 | Multi-speaker |
| HUT-Audio-Corpus-German | 326 | 44.1 | Multi-speaker |
| Blizzard-2013 | 319 | 44.1 | Single-speaker |
| M-AILABS | 79 | 16 | Both |
| Persian-text-to-speech | 30 | 22.05 | Single-speaker |
| **ArmanTTS** | 22.05 | 9 | Single-speaker |

There have been a number of works in the field of TTS, each with their own collection of datasets [3]–[5], [7], [10]–[12]. In this section we examine previous researches done in the field of TTS. Text to speech models are comprised of Acoustic models and vocoders. Acoustic models convert text to spectrograms and the vocoders convert spectrograms to waveforms. We will inspect both techniques in more detail.

*A. Acoustic Models*

Acoustic models can be categorized as either SPSS or end-to-end. HMM[13] models and RNN based BiLSTM[14] are examples of SPSS. These models take linguistic features from the input and produce acoustic features in the output. End-to-end models on the other hand directly take the characters or phonemes as input and produce the acoustic features as output. Tacatron[1], TransformerTTS[2], and Fast Speech[15] are examples of the end-to-end model. The results of our experimentation with the Tacatron 2[9] model is reported in Experiment.

*B. Vocoders*

Vocoders are classified into four categories: autoregressive, flow-based, GAN-based and diffusion-based [16]. WaveNet [17] is a vocoder that uses dilated convolutions [18] to create autoregressive wave points. Flow-based models are a type of generative model in which invertible mappings are used for transforming the probabilistic density. Based on their type of transformation, flow-based models are either autoregressive transformers or bipartite transformers. Autoregressive transformations are more accurate while bipartite transformations have a simpler training process [16]. GAN-based models, however, consist of a generative part that generates synthetic data and a separator part to check the similarity of synthetic data to real data in the training phase [19]. The Mel-GAN[12], HiFi-GAN[3], GAN-TTS[4], and Wave-GAN[12] models are examples of GAN-based vocoders [16]. We experimented with the HiFi-GAN vocoder, which results are reported in Experiment. Finally, the diffusion-based model uses Denoising Diffusion Probabilistic Models (DDPM or Diffusion) [20] and converts acoustic features into waveforms [16].

*C. Datasets*

TTS datasets require samples that are noise-free and of good quality. This can make the data collection process difficult and costly. Because of this, TTS datasets are relatively scarce and do not exist for most languages. Although there have been researches trying to reduce the cost of preparing TTS dataset with solutions such as unsupervised learning, cross-linguistic transformation, removing noise and increasing the quality of environmental sounds, etc [16].

Persian-text-to-speech [8] is currently the only available dataset in Persian. Therefore, in order to expand TTS in the Persian language, we have collected a new dataset and presented it in this paper. There are various datasets for TTS in other languages, particularly in English which we will review below;

*1) LibriTTS Dataset*

LibriTTS[5] is a multi-speaker English dataset derived from the LibriSpeech corpus [21]. This dataset contains 582 hours of speech and 2456 speakers, with a sample rate of 24kHz. Each sample in the dataset contains one sentence, each including the original text and the normalized text of the samples.

*2) LJSpeech Dataset*

The LJSpeech dataset [6] is a single-speaker English dataset derived from the LibriVox books [22]. It contains about 24 hours of speech with a sample rate of 22.05 kHz.

*3) HiFi-TTS Dataset*

The HiFi-TTS dataset [7], is a high quality English dataset with 292 hours of speech and 10 speakers. The sample rate seen in this dataset is above 44.1 kHz.

*4) HUI-Audio-Corpus-German Dataset*

HUI-Audio-Corpus-German[23] is a high quality German dataset. It contains speech from 122 speakers for a sum of 326 hours. The sample rate of this dataset is 44.1 kHz.

*5) Blizzard-2013 Dataset*

The Blizzard-2013 dataset [10] resulted from a challenge presented in 2013 to generate artificial speech. It contains about 319 hours of single-speaker speech in Indian, with a sample rate of 44.1 kHz.

*6) M-AILABS Dataset*

M-AILABS[24] is a dataset for converting from speech to text, also involving automatic speech recognition. It contains 75 hours of speech with a sample rate of 16kHz. The samples in the dataset are mono-wave, containing female voice only, male voice only, and a combination of female and male voice. The samples consist of various Latin languages such as English, German, and Russian.

*7) Persian-Text-To-Speech Dataset*

The Persian-text-to-speech[8] dataset, is composed of samples collected from audiobook recordings, chosen based on the availability of the corresponding text. The sampling rate of the original sounds is 44.1kHz, which is reduced to 22.05kHz for the dataset. The number of audio channels of the samples was reduced to single channel files. This dataset contains 30 hours of single-speaker speech. Because of the long duration of the original audio file, shorter samples were extracted using a silence detector.

### III. DATASET STRUCTURE

To produce ArmanTTS Farsi text-to-speech dataset, we took the following steps;

*A. Persian Text*

To obtain the persian text of the audio files, we used OpenSubtitles[25]. Each sentence was separated into a separate audio file, resulting in 9 hours, 12 minutes, and 14 seconds of audio duration. Figure 2 shows the distribution of sentence length throughout the dataset. As you can see, the number of words in most samples is below 10. Figure 1 shows the duration histogram of the audio files. According to Figure 1, the duration of most audio files is close to 2.5 seconds.

*B. Mapping Phonemes and Letters*

In this dataset, the model input is the phoneme. Table III shows the mappings between different phonemes and certain persian words. Table III contains 29 phonemes; e, o, i, A, a and u are vowels and the rest are consonants. In the following, we see the mapping of some examples of sentences to phonemes.

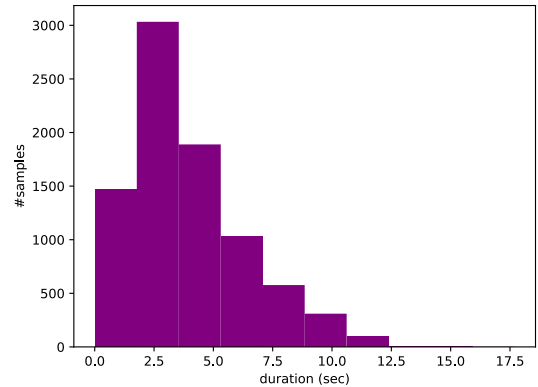

Figure 1. The histogram of duration time of the audios in the dataset

Example 1:

دو ساله که با هم صحبت نکرده‌اند

دُ سالِ کِ بآ هَم صُحَبَت نَکَردِنَّد

do sAle ke bA ham sohbat nakarde'and

Example 2:

این مرغ‌ها را از کجا گرفتی

ئین مُرغهآ رآ ئَز کُجآ گِرِفتی

'in morqhA rA 'az kojA gerefti

Example 3:

تو نمیدونی داری چیکار میکنی

تُ نِمیدونی دآری چیکآر میکُنی

to nemiduni dAri CikAr mikoni

Each example is written in three forms. For each example above, the first line is written in the common form of Persian text. In the second line, the sentence is written so that it can be mapped with phonemes, and in the third line, the sentence is written using phonemes. The third line is the input of the model in the ArmanTTS dataset.

*C. Normalizing the text*

In the ArmanTTS dataset, the input to the model is phonemes. Therefore, there is no process to normalize the text. Figure 4 shows the occurrence frequency of each phoneme in the dataset. According to Figure 4, vowel phonemes occurred more than consonant phonemes, which is natural.

*D. Audio and Speech*

The speech production of this dataset was done in a studio environment with a sampling rate of 22.05 kHz. The average signal to noise ratio is 25 dB. Also, this dataset is single-speaker. Figure 3 shows the signal-to-noise histogram of the dataset files. It can be seen that the signal-to-noise ratio of the majority of the audio files is above zero, which indicates that the dataset is acceptable in terms of the amount of noise.

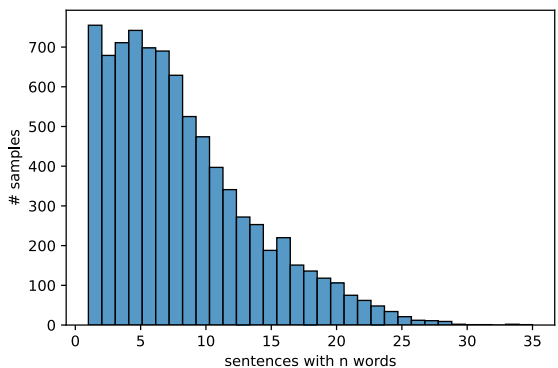

Figure 2. histogram of the sentence length of the samples in the data

## E. Categorization

The dataset consists of two separate sections, training and testing. The training set contains 8419 samples and the test set contains 30 samples.

## IV. EXPERIMENT

To evaluate the ArmanTTS dataset, we used the text-to-speech model consisting of the HiFi-GAN[3] vocoder and the Tacotron 2[9] acoustic model. We used MOS(mean opinion score) metrics for the evaluation. MOS is a metric used for evaluation of ground-truth audio, TTS models and vocoders. We evaluated 30 samples which results can be seen in Table II. 16 people evaluated the audio file and gave each one a score truth speech for 8 of them, the next 7 samples have ground-truth spectrogram and we only used vocoder model to evaluate the vocoder, and TTS model output for the remaining 15. MOS of the ground-truth speech was evaluated as 4.0, for the vocoder evaluated as 3.86 and for the TTS model was 2.98.

TABLE II. EVALUATION OF TTS MODEL ON ARMANTTS DATASET AND ITS COMPARISON WITH REAL EXAMPLES

| Type | Number of samples | Number of voters | MOS (0-5) |
| --- | --- | --- | --- |
| Ground-truth | 8 | 16 | 4.0 |
| Vocoder | 7 | 16 | 3.87 |
| TTS model prediction | 15 | 16 | 2.98 |

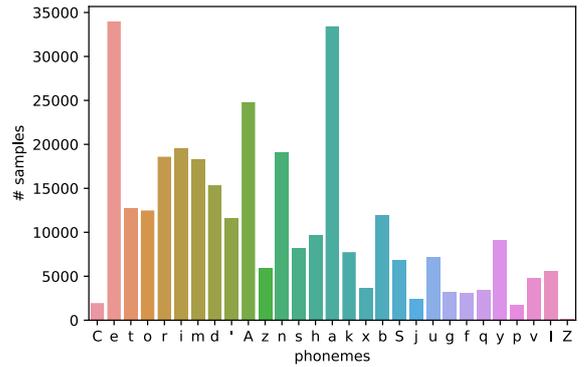

Figure 4. histogram of the number of occurrences of each phoneme in the data

TABLE III. PHONEME AND LETTER MAPPING OF PERSIAN TEXT

| phoneme | characters |
| --- | --- |
| C | چ |
| e | ِا |
| t | ت ط |
| o | اً |
| r | ر |
| i | ای |
| m | م |
| d | د |
| ' | ع ئ ء |
| A | آ |
| z | ز ظ ض ذ |
| n | ن |
| s | س ص ث |
| h | ه ح |
| a | اَ |
| k | ک |
| x | خ |
| b | ب |
| S | ش |
| j | ج |
| u | او |
| g | گ |
| f | ف |
| q | ق غ |
| Y | ی |
| p | پ |
| v | و |
| l | ل |
| z | ژ |

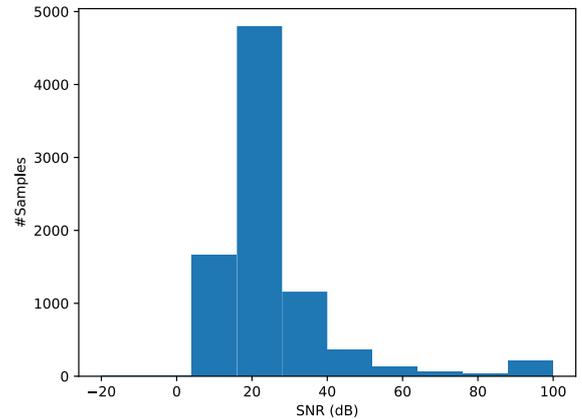

Figure 3. histogram of the signal-to-noise of each audio sample in the data

between 0 and 5. Out of these 30 samples, we used ground-

The results show that we can achieve acceptable performance with ArmanTTS dataset. However, we need larger amount of data to cover the problem domain appropriately to reach better performance.

## V. CONCLUSION

Due to the lack of datasets in the field of Persian text to speech conversion, in this paper we introduced the ArmanTTS dataset. It includes a single male speaker and contains 8449 Persian audio samples, resulting in 9 hours and 12 minutes of speech. Additionally, we evaluated the dataset with a base method. This method allowed us to model and test the dataset. The resulted evaluation obtained for the ground-truth voice, vocoder prediction with ground-truth acoustic features, and the TTS model voice, are respectively 4.0 MOS, 3.86 MOS and 2.98 MOS. Of course, there is still a lot of potential to expand in Persian text-to-speech conversion. Some of the measures that can be taken to expand this area is the collection of multilingual datasets, datasets with less noise, higher sampling rates and longer durations for the Persian language.

## ACKNOWLEDGEMENT

This dataset was collected with the aid of the Arman Rayan Sharif Company located in Iran. We thank this company for their full support in producing and publishing this dataset.